\documentclass[times, review, 10pt]{elsarticle}

\usepackage{threeparttable}
\usepackage{amsmath}
\usepackage{booktabs}
\usepackage{multirow}
\usepackage{amssymb}

\journal{Pattern Recognition}

\begin{document}

\begin{frontmatter}



\title {Boosting Overlapping Organoid Instance Segmentation Using Pseudo-Label Unmixing and Synthesis-Assisted Learning}  


\author[1]{Gui Huang}
\ead{huangg55@mail.sysu.edu.cn}

\author[1]{Kangyuan Zheng}
\ead{zhengky29@mail2.sysu.edu.cn}

\author[1]{Xuan Cai}
\ead{caix63@mail2.sysu.edu.cn}

\author[1]{Jiaqi Wang}
\ead{jackiewangsysu@gmail.com}

\author[1]{Jianjia Zhang}
\ead{zhangjj225@mail.sysu.edu.cn}

\author[2]{Kaida Ning}
\ead{ningkd@pcl.ac.cn}

\author[3]{Wenbo Wei}
\ead{wwb1004@hotmail.com}

\author[4]{Yujuan Zhu}
\ead{yujuanzhu@ntu.edu.cn}

\author{Jiong Zhang\corref{cor1}\fnref{5}}
\ead{jiong.zhang@ieee.org}

\author{Mengting Liu\corref{cor1}\fnref{1}}
\ead{liumt55@mail.sysu.edu.cn}

\cortext[cor1]{}

\cortext[1]{Corresponding author}

\affiliation[1]{organization={School of Biomedical Engineering, Sun Yat-sen University},
	addressline={No. 66 Gongchang Road, Guangming District}, 
	city={Shenzhen},
	postcode={518107}, 
	state={Guangdong},
	country={China}}

\affiliation[2]{organization={Peng Cheng Laboratory},
	addressline={Building 2, Xili University Town}, 
	city={Shenzhen},
	postcode={518055}, 
	state={Guangdong},
	country={China}}

\affiliation[3]{organization={The First Affiliated Hospital of Shenzhen University, Shenzhen Second People's Hospital},
	addressline={No. 3002, Yantian Road, Luohu District}, 
	city={Shenzhen},
	postcode={518035}, 
	state={Guangdong},
	country={China}}

\affiliation[4]{organization={The Research Center of Clinical Medicine, Affiliated Hospital of Nantong University, Medical School of Nantong University},
	addressline={No. 20 Xisi Road}, 
	city={Nantong},
	postcode={226001}, 
	state={Jiangsu},
	country={China}}

\affiliation[5]{organization={Cixi Institute of Biomedical Engineering, Ningbo Institute of Materials Technology and Engineering, Chinese Academy of Sciences},
	addressline={No. 99, Xuelin Road}, 
	city={Cixi},
	postcode={315300}, 
	state={Zhejiang},
	country={China}}

\begin{abstract}
	Organoids, sophisticated \textit{in vitro} models of human tissues, are crucial for medical research due to their ability to simulate organ functions and assess drug responses accurately. Accurate organoid instance segmentation is critical for quantifying their dynamic behaviors, yet remains profoundly limited by high-quality annotated datasets and pervasive overlap in microscopy imaging. While semi-supervised learning (SSL) offers a solution to alleviate reliance on scarce labeled data, conventional SSL frameworks suffer from biases induced by noisy pseudo-labels, particularly in overlapping regions. Synthesis-assisted SSL (SA-SSL) has been proposed for mitigating training biases in semi-supervised semantic segmentation. We present the first adaptation of SA-SSL to organoid instance segmentation and reveal that SA-SSL struggles to disentangle intertwined organoids, often misrepresenting overlapping instances as a single entity. To overcome this, we propose Pseudo-Label Unmixing (PLU), which identifies erroneous pseudo-labels for overlapping instances and then regenerates organoid labels through instance decomposition. For image synthesis, we apply a contour-based approach to synthesize organoid instances efficiently, particularly for overlapping cases. Instance-level augmentations (IA) on pseudo-labels before image synthesis further enhances the effect of synthetic data (SD). Rigorous experiments on two organoid datasets demonstrate our method's effectiveness, achieving performance comparable to fully supervised models using only 10\% labeled data, and state-of-the-art results. Ablation studies validate the contributions of PLU, contour-based synthesis, and augmentation-aware training. By addressing overlap at both pseudo-label and synthesis levels, our work advances scalable, label-efficient organoid analysis, unlocking new potential for high-throughput applications in precision medicine.
\end{abstract}

\begin{highlights}
	\item Proposes Pseudo-Label Unmixing to fix semi-supervised errors via mask decomposition.
	\item Employs contour-based synthesis preserving organoid intersectional relationships.
	\item Optimizes synthetic-data integration in the semi-supervised learning framework.
	\item Introduces novel metric for evaluating synthetic data distribution fidelity.
\end{highlights}


\end{frontmatter}



\section{Introduction}
An organoid is a precisely engineered, three-dimensional (3D) tissue construct derived from pluripotent, fetal, or adult stem cells sourced from either healthy individuals or patients \cite{ref1}. These structures meticulously mimic the functional, structural, and biological intricacies of an organ, serving as an innovative model that integrates patient-specific characteristics with the capability to replicate \textit{in vivo} tissue architectures and functionalities in a controlled, \textit{in vitro} environment. Consequently, organoids have emerged as invaluable tools in a broad spectrum of applications, including drug discovery, tailored companion diagnostics, and advanced cell therapies \cite{ref2}. Regular microscopic imaging and observation of organoids are crucial for elucidating their morphological and growth characteristics, thereby playing a fundamental role in advancing organoid research \cite{ref3}. To transform these visual observations into quantifiable analysis, precise quantification of organoid morphology and behavior is vital. Instance segmentation—the task of delineating individual organoids with pixel-level masks—serves as the foundation for such analyses \cite{ref4}. However, the dense and rapidly evolving patterns of organoids inherently lead to frequent overlap, where multiple instances intertwine spatially, thus obscuring boundaries and confounding automated segmentation algorithms.

In detail, this complexity of organoid instance segmentation stems from inherent biological and technical factors. Organoids, by design, are cultured in a three-dimensional volume. When imaged via conventional microscopy, this volumetric structure is projected onto a two-dimensional plane, collapsing organoids at varying depths into overlapping clusters. Consequently, organoids positioned at different layers within the organoid appear superimposed in the image, creating ambiguities that confound segmentation algorithms. Compounding this issue is the dense arrangement intrinsic to organoids, resulting in frequent physical contact and partial overlap between neighboring instances. These biological characteristics are further exacerbated by limitations in sample preparation and imaging technologies. For instance, planar microscopy techniques, widely used for high-throughput screening, lack the optical sectioning capabilities required to resolve axial depth \cite{ref5}, while mechanical compression during sample mounting may artificially induce multi-layered organoid stacking \cite{ref6}. Together, these factors generate images where overlapping boundaries and regions dominate, rendering traditional instance segmentation approaches ineffective.

Manual instance segmentation of these images is both tedious and time-consuming \cite{ref7},\cite{ref8}, requiring domain expertise to navigate ambiguities in overlapping structures. To alleviate this burden, deep learning techniques have emerged as potent tools for automating organoid-related tasks \cite{ref9}. Current deep learning-based models for instance segmentation, such as Mask R-CNN, rely heavily on large-scale annotated datasets to achieve robust performance \cite{ref10}. However, existing publicly accessible organoid instance segmentation datasets remain limited in scale. Semi-supervised learning (SSL) has emerged as an effective strategy to reduce the reliance on scarce labeled data by leveraging unlabeled data through pseudo-label supervision. Since initial pseudo-labels often exhibit suboptimal quality, they require refinement based on prior knowledge to prevent training biases, necessitating careful evaluation and optimization processes. Furthermore, we observe that conventional -SSL struggles with organoid overlap, often misrepresenting multiple organoids as a single entity, distorting the spatial relationships critical for accurate instance segmentation. 

Recent advances in synthesis-assisted SSL (SA-SSL) aim to circumvent explicit pseudo-label evaluation and refinement by generating synthetic images aligned with pseudo-labels as supplement training data \cite{ref11}. Image synthesis is a key process in SA-SSL and often involves image-to-image transformations using inputs like texts \cite{ref12}, masks \cite{ref13}, sketches \cite{ref14}, contours \cite{ref15}, or other image modalities \cite{ref16}. In organoid image synthesis, this task is commonly framed as semantic mask-to-image translation, where a semantic mask is utilized to generate an image. However, when synthesizing overlapping structures, multiple semantic masks generated via distributing overlapping instances into different layers are required to assign the shared intersecting regions to all participating instances, ensuring that each instance's mask retains complete spatial information. The ‌increasing computational costs‌ motivate alternative approaches. Contour-based representations offer a promising solution by encoding instance intersections as limited number sparse intersection points in contour images. Assigning these points to a single instance results in minimal loss of instance information. Therefore, it presents a potentially efficient and viable option for synthesizing overlapping structures using only a single input.

In this study, we confront the challenges when introducing the synthesis-assisted learning into the semi-supervised organoid instance segmentation. Our approach explicitly targets the root causes of segmentation errors—overlap, through a correction novel strategy grounded in mask decomposition (Pseudo-Label Unmixing, PLU). Given the scarcity of high-quality labeled organoid datasets, we manually labeled images documented by Kassis \textit{et al}. \cite{ref17} to create a diverse dataset covering various scenarios, especially overlap. We enhance model robustness by generating synthetic images from contour-based representations that preserve intersectional relationships, narrowing the gap between synthetic and real data distributions. By augmenting pseudo-labels via instance-level augmentations (IA) before image synthesis, we increase synthetic data diversity. We introduce a quantitative metric to evaluate synthetic-real distribution alignment and demonstrate through experiments that augmentation strategies balancing diversity and distribution fidelity improve model performance. This establishes a paradigm for leveraging augmented synthetic data to improve label efficiency. By harmonizing pseudo-labels correction, synthesis fidelity, and augmentation-aware training, our approach unlocks new possibilities for accurate, scalable analysis of complex organoid systems, advancing their utility in high-stakes applications such as drug screening and disease modeling.

In summary, our contributions encompass:
\begin{itemize}
	\item {We present the first identification of a fundamental limitation inherent to SSL for resolving overlapping organoid structures. To address this issue, we propose a novel two-stage strategy that first detects erroneous pseudo-labels arising from SSL's failure cases, then performs mask decomposition to achieve accurate label correction.}
	\item {To our knowledge, this is the first time a contour-based image synthesis approach has been employed to tackle complex organoid scenarios with overlap, accompanied by a rigorous evaluation of computational efficiency. }
	\item {We present the first adaptation of SA-SSL to organoid instance segmentation, while simultaneously performing a comprehensive investigation into the optimal integration strategies for incorporating synthetic data (SD) within the SSL framework through systematically evaluating impacts of IA on model performance to advance SD utilization.}
	\item {We have conducted extensive experiments on organoid datasets, demonstrating that our methods significantly outperform existing schemes for semi-supervised segmentation. This underscores the remarkable effectiveness of our approach in addressing organoid segmentation tasks.}
\end{itemize}

\section{Related work}
\subsection{Organoid Instance Segmentation}
Instance segmentation plays a vital role in dynamic organoid analysis but presents considerable challenges due to the extensive diversity in organoid morphology, their low contrast against backgrounds, and the frequent occurrence of overlap. To tackle these challenges, recent studies have incorporated deep learning methods for automated organoid segmentation. Various network architectures, encompassing both CNNs \cite{ref18} and transformer-based models \cite{ref19}, have been applied in this field. Although transformer-based frameworks show great segmentation accuracies, their adoption remains constrained by the lack of extensive large-scale datasets. Consequently, CNNs, particularly U-Net \cite{ref20} and Mask-RCNN \cite{ref21}, have emerged as prevalent choices for organoid instance segmentation. Notably, Mask-RCNN has demonstrated superior performance in handling overlapping objects compared to U-Net \cite{ref22}. However, most research endeavors follow a supervised learning (SL) paradigm to train models. While these models exhibit commendable performance, their full potential may remain untapped when confronted with limited labeled data. 
In this study, we aim to develop an advanced framework to improve the segmentation performance of Mask-RCNN models, with a particular focus on addressing the challenges associated with limited labeled data.

\subsection{Semi-Supervised Instance Segmentation}
Collecting pixel-level annotations for instance segmentation, particularly in the context of organoids, is a labor-intensive and expensive process. Consequently, SSL methods that leverage both labeled and unlabeled data have gained significant attention as a means to train segmentation models efficiently. Nevertheless, in comparison to semi-supervised object detection and semantic segmentation, semi-supervised instance segmentation has received limited attention and produced only a few successful endeavors. Among them, ‌Noisy Boundaries‌ \cite{ref23} stands out as a pioneering effort, improving model performance by enabling the student network to utilize low-resolution features to learn a noise-tolerant mask head specifically for boundary prediction. ‌In a subsequent study, Polite Teacher \cite{ref24} was introduced, which incorporates mask scoring to refine pseudo-mask thresholding, thereby enhancing its overall robustness. ‌PAIS \cite{ref25} was later introduced, focusing on leveraging pseudo-labels with lower confidence levels to unlock their potential. ‌Guided Distillation \cite{ref26} developed thereafter, introduced a unique form of knowledge distillation and an improved burn-in stage prior to initiating the main semi-supervised training loop. Notably, these efforts predominantly adopt the Teacher-Student paradigm, focusing on advancing network architectures or refining pseudo-labels.

In this work, we adopt a semi-supervised instance segmentation framework under the teacher-student paradigm, enhanced by synthesis-assisted learning. Unlike existing methods, this framework eliminates complex loss designs and regularization requirements while maintaining robustness against low-quality pseudo-labels. The effectiveness of this approach in organoid instance segmentation is evaluated on organoid datasets, demonstrating its potential in improving segmentation accuracy and robustness.

\subsection{GANs-Based Organoid Image Synthesis}
GANs have been widely adopted in biomedical image synthesis due to their ability to generate high-quality and realistic images \cite{ref27},\cite{ref28},\cite{ref29}. In organoid image synthesis, several studies have explored the use of GANs to address challenges such as limited data availability and the need for high-resolution synthetic images for downstream tasks. For instance, Du \textit{et al}. \cite{ref30} explored the capability of GANs in synthesizing different organoid types using sketches as inputs, highlighting its effectiveness in generating organoid images of biological relevance. Hradecka \textit{et al}. \cite{ref31} utilized semantic masks generated by traditional image processing algorithms as inputs for GANs, producing synthetic images that were subsequently used to train a deep learning network for accurate segmentation. Huang \textit{et al}. \cite{ref32} investigated the efficiency of GANs in synthesizing fluorescent images from bright-field images, demonstrating its potential as a virtual screening method for systematically evaluating organoid morphological heterogeneity in drug responses. 
However, current approaches primarily focus on well-separated organoids while neglecting prevalent overlap challenges in experimental settings. To address this gap, our study presents a novel methodology that integrates contours into the image synthesis process, harnessing their intrinsic capability to effectively manage organoid overlap. Furthermore, we propose IA applied to pseudo-labels prior to synthesis, which enhance the effectiveness of synthetic images, ultimately leading to improved segmentation performance. 

\section{Method}
In this section, we first introduce the development of datasets, including the selection and annotation of the dataset, and then introduce the problem of semi-supervised organoid instance segmentation. After that, we present the process of pseudo-label generation and elaborate the idea of PLU. Subsequently, we introduce the key details of pseudo-label guided image synthesis and highlight the effectiveness of the contour-based approach in organoid image synthesis, especially in tackling overlap. Leveraging the synthetic images, we comprehensively present the proposed SA-SSL framework.

\subsection{Dataset Development}
We evaluated our method on two organoid datasets: OrganoSegment and a newly constructed dataset, M-OrgaQuant. OrganoSegment \cite{ref9} offers a relatively large scale (231 images, 15,515 instances), making it suitable for semi-supervised learning. However, since overlapping organoids in this dataset had shared regions assigned to only one instance, we re-annotated such cases for more accurate ground truth. To complement this, we further developed M-OrgaQuant from the extensive OrgaQuant image collection originally curated by Kassis \textit{et al}. \cite{ref17}, comprising 1,752 images and 20,499 instances. Unlike the original OrgaQuant, which was tailored for detection tasks, M-OrgaQuant was specifically adapted for segmentation and clearly differentiated by name. The images span a wide range of real-world challenges—including occlusions, overlapping structures, blurred spheroids, and adverse lighting conditions—providing a comprehensive representation that enhances model robustness and generalization. Some other well-known open-source datasets, such as OrganoID (94 images, 5,108 instances) \cite{ref33} and OrgaExtractor (30 images, 2,202 instances) \cite{ref34}, were excluded from our study due to their limited sample sizes.

Segmentation annotation was performed by an interactive semi-automatic annotation tool based on Segment Anything Model (SAM) and can be accessed on GitHub at https://github.com/yatengLG/ISAT\_with\_segment\_anything. In this process, each image was independently labeled by two different experts. If the Intersection over Union (IoU) between the results of the same instance was less than 80\%, the image was then presented to a third expert for further annotation. The final segmentation for each image was chosen based on an aggregate decision, requiring at least 70\% agreement among all participating experts. To ensure the quality of the annotations, detailed instructions and examples were provided to the experts, who were required to pass a quality test before undertaking the task.

\subsection{Problem Definition}
The task of this research is formulated as follows. A dataset \textit{D}, comprising a limited number \textit{n} of labeled images and an abundant supply \textit{m} of unlabeled images is utilized for instance segmentation model training. Typically, $n\ll m$. SSL endeavors to leverage the vast amount of unlabeled data to achieve improved performance, aiming to match or exceed the results of models trained on fully labeled datasets. Nonetheless, as shown in 
ure 1, pseudo-labels generated from unlabeled data contain missing or incorrect predictions, particularly in challenging scenarios involving severe overlap—defined as cases where ‌$>1/3$ of an organoid's own area is occupied by overlapping regions with neighboring structures (highlighted by red rectangles). This results in a mismatch between images and their corresponding labels, which can introduce training bias and significantly impede performance improvement. To address this issue, we introduce pseudo-label guided image synthesis as an efficient strategy to enhance image-label alignment and increase the quantity of overlapping structures. Additionally, we propose PLU to rectify erroneous predictions for overlapping structures. Furthermore, we apply IA to pseudo-labels prior to image synthesis to amplify the effect of synthetic images through enhanced data diversity. Our goal is to reduce training bias caused by pseudo-labels in SA-SSL, thereby achieving superior model performance, particularly for overlapping structures.

\begin{figure}
	\centering
	\includegraphics[width=0.9\linewidth]{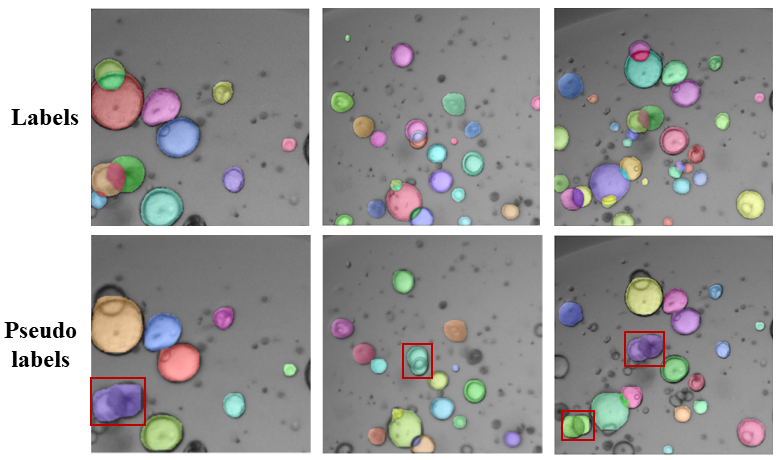}
	\caption{Labels and pseudo labels of organoids. Incorrect pseudo labels of overlapping structures are highlighted by red rectangles.}
	\label{fig1}
\end{figure}

\subsection{Pseudo-Label Generation and Correction}
As illustrated in Figure 2, our framework is built on Mask R-CNN, a two-stage instance segmentation architecture that first detects instances then segments them. In SSL, pseudo-label generation of this pipeline involves two-step thresholding of the raw inference masks of teacher model \textit{T}. First, bounding box confidence thresholding filters low-quality boxes with a confidence threshold. Second, pixel-level probability thresholding with threshold $\theta_{p}$ precisely separates foreground/background pixels for accurate segmentations:
\begin{equation}
	M_{ij} =
	\begin{cases}
		1, & \text{if } P_{ij} \geq \theta_{p} \\
		0, & \text{if } P_{ij} < \theta_{p}
	\end{cases}
	\label{eq}
\end{equation}
where $M_{ij}$ and $P_{ij}$ represent the mask value and probability for the \textit{j}-th pixel in the \textit{i}-th mask, respectively. In accordance with established practices, we set the bounding box-based threshold to 0.7 and the pixel-based threshold to 0.5. The generated pseudo-labels are typically employed directly in the subsequent image synthesis process. Nevertheless, as illustrated in Figure 1, these pseudo-labels often contain errors, particularly when organoid overlap cause multiple instances to be incorrectly merged. These errors, if uncorrected, lead to a reduced representation of overlapping organoids in the synthetic images, which in turn hinder the model's ability to effectively learn and process such cases.

To address this issue, we propose the PLU module integrated with Mask R-CNN. This module first detects erroneous pseudo-labels in overlapping structures, then reconstructs accurate masks through mask decomposition. This decomposition process predicts multiple masks using the feature map from erroneous cases, drawing inspiration from prior work on handling missing detections in crowded scenes \cite{ref35}. As shown in Figure 2, the modified pipeline starts with a Region Proposal Network (RPN) that generates candidate instance locations in the form of bounding boxes (\textit{i}.\textit{e}. proposals), guided by predefined anchor sizes. These proposals are then processed by a Feature Pyramid Network (FPN), which extracts multi-scale features to accommodate variant object sizes. Region of Interest (RoI) Align Module crops and resizes these features, enabling parallel processing through detection and segmentation branches. The detection branch consists of a classification head for foreground/background probability predictions and a bounding box regression head for coordinate refinement. Simultaneously, the segmentation branch employs convolutional layers followed by a deconvolution layer to generate pixel-wise masks for each RoI. The predicted masks are subsequently passed to an overlapping judgement branch (\text{Overlap\_judge\_head}), which identifies incorrect segmentations caused by overlapping instances. When the overlap judgment branch detects an erroneous predicted mask, where multiple overlapped organoids are mistakenly merged into a single entity, it triggers the overlapping decomposition branch to correct it by reconstructing accurate masks of the overlapped organoids from the corresponding feature map.

As illustrated in Figure 2, the entire model incorporates six loss functions ($L_{cls}$, $L_{reg}$, $L_{seg}$, $L_{\mathit{O}\_\mathit{cls}}$,  $L_{\textit{i}\_\textit{count}}$ and $L_{\textit{i}\_\text{IoU}}$) for SL. The first three losses are inherited from Mask R-CNN, handling classification, bounding box regression, and instance mask prediction, respectively. The remaining three losses are specifically designed to detect erroneous pseudo-labels and accurately decompose them. Detailed information of these losses are as follows:

\begin{figure*}
	\centering
	\includegraphics[width=0.9\linewidth]{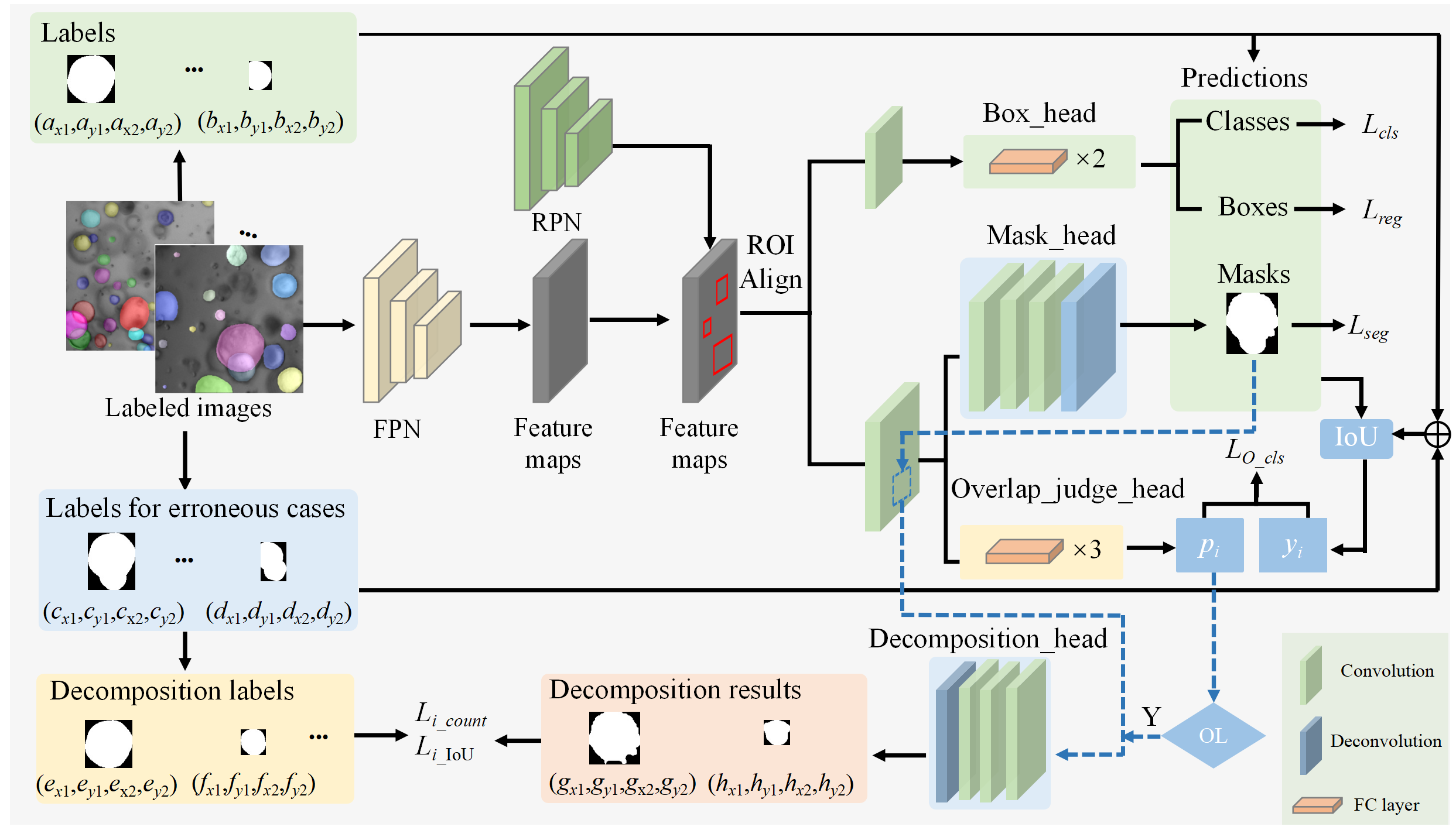}
	\caption{Illustration of pseudo-label correction. An overlapping judgment branch detects erroneous pseudo-labels, which are subsequently rectified through a decomposition branch to obtain a corrected version. FC: Fully connected, $p_{i}$: Model-predicted overlap judgments, $y_{i}$: Ground-truth overlap judgments.}
	\label{fig2}
\end{figure*}

1) $L_{cls}$ represents the focal loss for instance proposal classification: determining whether a proposed candidate region is foreground or background. It is an adaptive loss function designed to address class imbalance—particularly when the foreground-to-background ratio is very small—by dynamically adjusting the cross-entropy loss according to the classification difficulty. The complete formulation is:

\begin{equation}
	L_{cls}=-\alpha(
	\begin{array}
		{c}{1-p_{t}}
	\end{array})^{\gamma}\log(p_{t})
	\label{eq}
\end{equation}
where $\alpha$ is the class-balancing weight for foreground/background, $p_{t}$ denotes the foreground probability after sigmoid activation and $\gamma$ focuses learning on hard-to-classify samples.

2) $L_{reg}$ denotes the smooth-L1 loss for bounding box regression. It is a hybrid loss combines $\mathit{L}_{1}$ and $\mathit{L}_{2}$ advantages to reduce sensitivity to outliers, allowing robust bounding box coordinate prediction:

\begin{equation}
	L_{reg}(t,\nu)=\sum_{i\in\{x,y,w,h\}}
	\begin{cases}
		0.5\times(t_i-\nu_i)^2, & \text{if } |t_i-\nu_i|<1 \\
		|t_i-\nu_i|-0.5, & \text{otherwise}
	\end{cases}
	\label{eq}
\end{equation}
where $t_{i}$ and $v_{i}$ are the bounding box offsets of ground-truth and predicted results, respectively.

3) $L_{seg}$ signifies the pixel-wise cross-entropy loss for segmentation. It minimizes per-pixel classification error within each RoI to optimize instance mask prediction:
\begin{equation}
	L_{seg}=-\frac{1}{N}\sum_{i=1}^{N}\sum_{c=0}^{1}y_{i,c}\log p_{i,c}
	\label{eq}
\end{equation}
where $N$ is the total number of pixels, $c\in\{0,1\}$ are the class indices (0 for background, 1 for foreground), for a given pixel $i$, $y_{i,c}$ indicates its one-hot encoded ground truth label (1 if pixel \textit{i} belongs to class $c$, 0 otherwise) and $p_{i,c}$ represents its predicted probability for pixel $i$ being class $c$.

4) $L_{\mathit{O}\_\mathit{cls}}$ is the binary cross-entropy loss for Pseudo-Label Classification. It is a specialized loss function for distinguishing correct and erroneous pseudo-labels in overlapping structures, defined as:
\begin{equation}
	L_{\mathit{O}\_\mathit{cls}}=-\frac{1}{n}\sum_{i=1}^{n}\left[y_{i}\log(p_{i})+(1-y_{i})\log(p_{i})\right]
	\label{eq}
\end{equation}
where $n$ is the number of samples, $y_{i}$ indicates the ground-truth judgment (1 for correct pseudo-labels, 0 for erroneous ones) and $p_{i}$ is the predicted probability that the $i$-th pseudo-label is correct. The ground-truth judgements are derived by treating all manually segmented instances as correct examples, while erroneous samples are created by merging overlapping instances and combining their masks. The IoU scores of predicted masks against both the correct masks ($\text{IoU}_{c}$) and erroneous ones ($\text{IoU}_{e}$) are compared to determine the ground-truth judgments as:
\begin{equation}
	\begin{cases}
		1, & \mathrm{IoU}_c>\mathrm{IoU}_e \\
		0, & \mathrm{otherwise} 
	\end{cases}
	\label{eq}
\end{equation}

5) $L_{\textit{i}\_\textit{count}}$ denotes binary cross-entropy loss for instance counting in mask decomposition. This loss function enforces accurate instance counting in mask decomposition tasks, defined as:
\begin{equation}
	L_{\textit{i}\_\textit{count}}=-\frac{1}{K}\sum_{i=1}^{K}\left[k\log(e_{i})+(K-k)\log(1-e_{i})\right]
	\label{eq}
\end{equation}
where $K$ is the maximum possible instances, typically set to 5 as a default for balanced computational cost and granularity, $k$ denotes the ground-truth mask count and $e_{i}$ represents the predicted probability that the \textit{i}-th potential instance truly exists.

6) $L_{\textit{i}\_\text{IoU}}$ presents an IoU-based loss for decomposed mask alignment, computed after ‌optimal bipartite matching‌ through Hungarian algorithm \cite{ref36} which searches for best matched ground truth instance by IoU as cost. It evaluates spatial correspondence between all predicted and ground-truth mask pairs $\left\{\hat{Y}_{i},Y_{i}\right\}_{i=1}^{m}$ in overlapping regions, optimizing both alignment precision and instance separation in decomposed mask prediction. $L_{\textit{i}\_\text{IoU}}$ is calculated as:
\begin{equation}
	L_{i_{-}\mathrm{IoU}}=\frac{1}{m}\sum_{i=1}^{m}(1-\frac{\hat{Y}_{i}\bigcap Y_{i}}{\hat{Y}_{i}\bigcup Y_{i}})
	\label{eq}
\end{equation}
where $m$ is the total number of ground-truth mask pairs.The total training loss function is a weighted sum:
\begin{equation}
	L_{SL}=L_{cls}+L_{reg}+L_{seg}+\alpha L_{\mathit{O}\_\mathit{cls}}+\beta L_{\textit{i}\_\textit{count}}+\gamma L_{\textit{i}\_\text{IoU}}
	\label{eq}
\end{equation}
where $\alpha$, $\beta$and $\gamma$are trainable trade-off coefficients. Following the standard Mask R-CNN implementation, the default coefficients for the first three losses are uniformly set to 1.0. This equal-weight configuration has been empirically shown to provide balanced gradient contributions across tasks during multi-task optimization.

\subsection{Pseudo-Label Guided Image Synthesis.}
Pseudo-label guided image synthesis is typically regarded as an image-to-image translation process, where a trained generator \textit{G} synthesizes an image from a given semantic mask. However, when overlap exist in an image, a single semantic mask can no longer accurately represent all instance information, as the assignment of intersecting regions to a single instance inherently causes the loss of critical details, such as boundaries and spatial relationships between instances.

Motivated by Du \textit{et al}.'s approach to organoid image generation based on sketches \cite{ref30}, we introduce contours into pseudo-label guided image synthesis as a more efficient solution for generating organoid overlap. In this way, image synthesis can be formulated as:
\begin{equation}
	X_i^*=G^{^{\prime}}(S_i)
	\label{eq}
\end{equation}
where $G^{^{\prime}}$ is the generator capable of generating images from contour images, and $S_i$ is a given contour image. Compared to image generation based on masks, generating images from contours offers data efficiency, better generalization, and computational ease. Contours preserve topological clarity and allow flexible editing (adjust shapes without pixel-level masks), enhancing control in tasks of organoid imaging. The process of our proposed method is illustrated in Figure 3. It encompasses three main steps: contour generation, generator training, and image synthesis.

\textbf{Contour Image Generation}. Organoid contour images were generated from the contours of the instance segmentation masks. Additionally, we precisely extract the image features from the organoid instances to evaluate their transparency Tins and the degree of focus Fins in microscopic imaging. Based on these evaluations, we are able to categorize organoid instances into four distinct groups. This allows us to incorporate the transparency and the degree of focus of organoids into the generated contours, enhancing the distributional similarity between synthetic and real images. Specifically, we follow the method proposed by Bremer \textit{et al}. \cite{ref21} to quantify transparency by calculating the mean pixel value of each organoid instance and determine the degree of focus using the variance of the Laplacian gradient:

\begin{equation}
	T_{ins}=\frac{1}{N_{ins}}\sum_{i=1}^{N_{ins}}p_{i}
	\label{eq}
\end{equation}

\begin{equation}
	F_{ins}=\frac{1}{N_{ins}}\sum_{i=1}^{N_{ins}}(L_{i}-\overline{L})
	\label{eq}
\end{equation}
where $N_{ins}$ is the number of pixels in the organoid instance, $L_i$ is the Laplacian gradient of pixel \textit{i}, and $\overline{L}$ is the mean Laplacian gradient. The results are then utilized for categorization based on thresholds determined by analyzing the distributions of Tins and Fins across labeled samples.

\textbf{Generator Training.} We use the generated contour images to train a pix2pixHD \cite{ref37} model within a generative adversarial framework, as illustrated in Figure 3. A generator   is employed to produce images from contours. Meanwhile, discriminators ($\textit{D}_{1}$, $\textit{D}_{2}$ and $\textit{D}_{3}$) are utilized to differentiate between the real images and the generated images at three different scales. These scales are typically created by down-sampling the original images to lower resolutions (\textit{e}.\textit{g}., full, half, and quarter resolutions). During training,   strives to generate images that are as realistic as possible to cheat the discriminators ($\mathit{D_{i}}$), while each discriminator endeavors to accurately distinguish the real images from the fake ones. This can be formulated as a minmax game:
\begin{equation}
	\min_{G^{^{\prime}}}\left(\left(\max_{D_1,D_2,D_3}\sum_{i=1}^3L_{\mathrm{GAN}}(G^{^{\prime}},D_i)+\lambda\sum_{i=1}^3L_{FM}(G^{^{\prime}},D_i)\right)\right)
	\label{eq}
\end{equation}
where $L_{\mathrm{GAN}}(G^{^{\prime}},D_i)$ is the GAN loss and $L_{\mathrm{FM}}(G^{^{\prime}},D_i)$ is the feature matching loss. These losses are combined into the loss function using a weighting parameter $\lambda$, which controls the importance of the two terms. The adversarial training process iteratively improves the performance of both $G^{^{\prime}}$ and $D_i$, ultimately yielding a trained  $G^{^{\prime}}$ capable of producing high-quality images.

\begin{figure}
	\centering
	\includegraphics[width=0.9\linewidth]{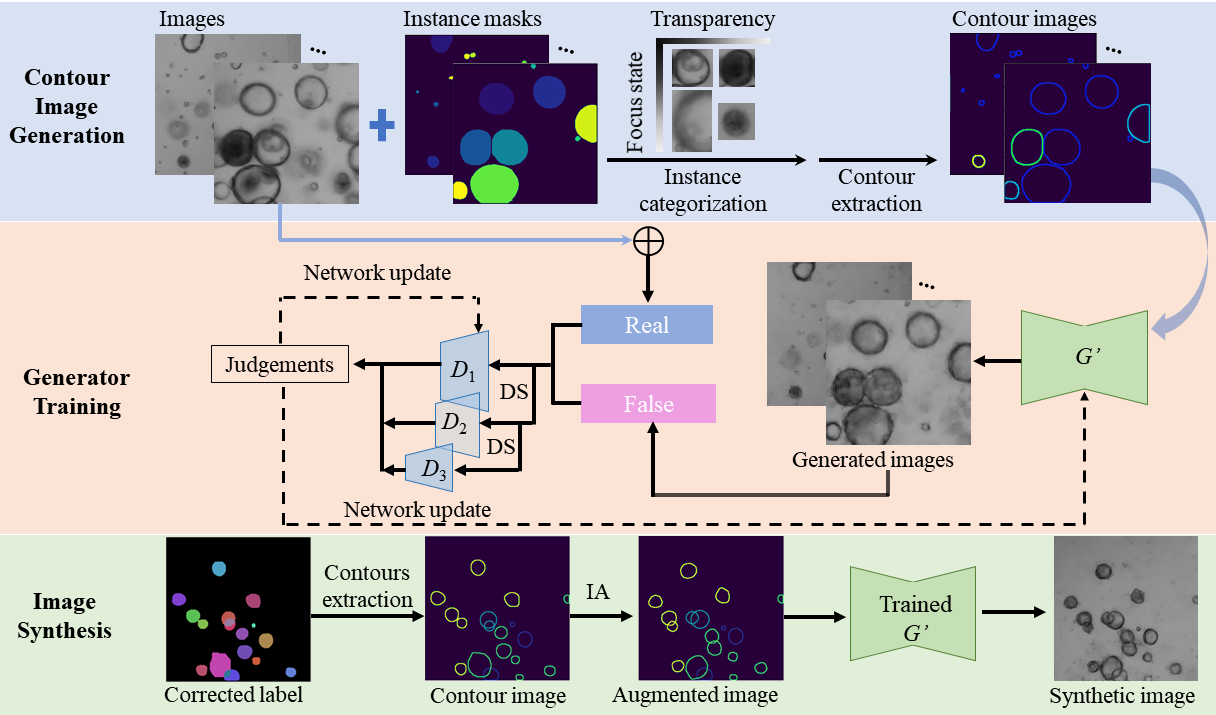}
	\caption{Workflow of our pseudo-label guided Image synthesis. Labeled data is categorized into four groups based on organoid transparency and focus state. Contour images of categorized organoids are generated and integrated into the generative adversarial training of the synthesis model, along with corresponding images. The modified model uses augmented images from corrected pseudo-labels to generate images. Here, instance masks are color-coded to distinguish individual organoids, while contour images use separate color schemes to represent different categories. DS: Down-sampling, IA: Instance-level Augmentations.}
	\label{fig3}
\end{figure}

\textbf{Image Synthesis.} The erroneous pseudo-labels were first rectified using the strategy outlined in the Pseudo-Label Generation and Correction section. The corrected segmentation masks were then transformed into contours via the contour generation process, which subsequently served as the basis for synthesizing the fake images. To ensure sufficient diversity in synthetic images, various augmentation techniques such as translation, rotation, and scaling, are applied to the contour images either individually or in combination. However, these instance-level augmentations might alter the spatial and structural properties of the contour images, leading to a distribution shift between synthetic and real images, which can negatively impact the effectiveness of synthetic images. Therefore, analyzing the impact of augmentation techniques on model performance is essential for optimizing augmentation strategies. To achieve this, we propose using Fréchet Inception Distance (FID) \cite{ref38} to evaluate the similarity of the distribution between synthetic and real images. This is typically accomplished by using features of real and generated images extracted from a pre-trained Inception network through the following formula:

\begin{equation}
	\mathrm{FID}((\mu_{\mathrm{r}},\sigma_{r}),(\mu_{\mathrm{g}},\sigma_{\mathrm{g}}))=
	\|\mu_{\mathrm{r}}-\mu_{\mathrm{g}}\|_2^2 + \mathrm{Tr}(\sigma_{r}+\sigma_{\mathrm{g}}-2\sqrt{\sigma_{r}\sigma_{\mathrm{g}}}).
	\label{eq}
\end{equation}

where $\mu_{\mathrm{r}}$ and $\mu_{\mathrm{g}}$ represent the mean feature vectors of real and generated images, respectively, while $\left\|\mu_{\mathrm{r}}-\mu_{\mathrm{g}}\right\|_{2}^{2}$ denotes the squared Euclidean distance between the mean feature vectors. $\sigma_{r}$ and $\sigma_{g}$ are the covariance matrices of real and generated images, respectively, and $\sqrt{\sigma_r\sigma_g}$ represents the matrix square root of the product of the covariance matrices. Tr denotes the trace of a matrix, which is the sum of its diagonal elements.

\subsection{SA-SSL}
In semi-supervised segmentation, pseudo-labels are commonly employed but often contain errors that need refinement, inevitably adding extra labor and uncertainty. SA-SSL circumvents these challenges. In this framework, pseudo-labels guide the training of both real and synthesized images. Although pseudo-labels for real images are imperfect, those for synthesized images match perfectly. This is because pseudo-label guided image synthesis generates new images that closely align with the pseudo-labels, creating additional reliable image-label pairs for improved instance segmentation training. In our study, pseudo-label-real-image pairs and pseudo-label-synthetic-image pairs are equally weighted during training.

We adopt the Mask R-CNN \cite{ref39} architecture as the backbone for instance segmentation. Figure 4 illustrates our framework, which integrates pseudo-label decomposition with SA-SSL segmentation. Teacher model \textit{T} is first trained on limited labeled images through SL. The trained teacher model is then used to generate reliable predictions for unlabeled images after weak augmentations (\textit{e}.\textit{g}., scaling and horizontal flipping). These predictions undergo a pseudo-label generation process to create raw pseudo-labels, which are then subjected to a PLU process. The corrected pseudo-labels and their corresponding synthetic images form the SD. Three sets of images were utilized to train the student model \textit{S}: real images with ground truth labels, real images with pseudo-labels, and synthesized images with pseudo-labels. During student model training, strong augmentations (\textit{e}.\textit{g}., color jittering, grayscale conversion, Gaussian blur, and cutout) are applied to enhance the generalization ability of \textit{S}. The loss computation for real data, pseudo data (PD), and SD follows the same approach as SL. The total loss is a weighted average of the three losses:
\begin{equation}
	L_{SA-SSL}=L_{real}+\lambda(L_{pseudo}+L_{synthetic})
	\label{eq}
\end{equation}
where $\lambda$  is a trade-off coefficient. Finally, an Exponential Moving Average (EMA) of the student's weights is periodically updated to the teacher model. This allows the teacher model to acquire the most recent and refined knowledge from the student, consequently enhancing its generalization capability and performance, and facilitating a continuous learning cycle.

\begin{figure}
	\centering
	\includegraphics[width=0.9\linewidth]{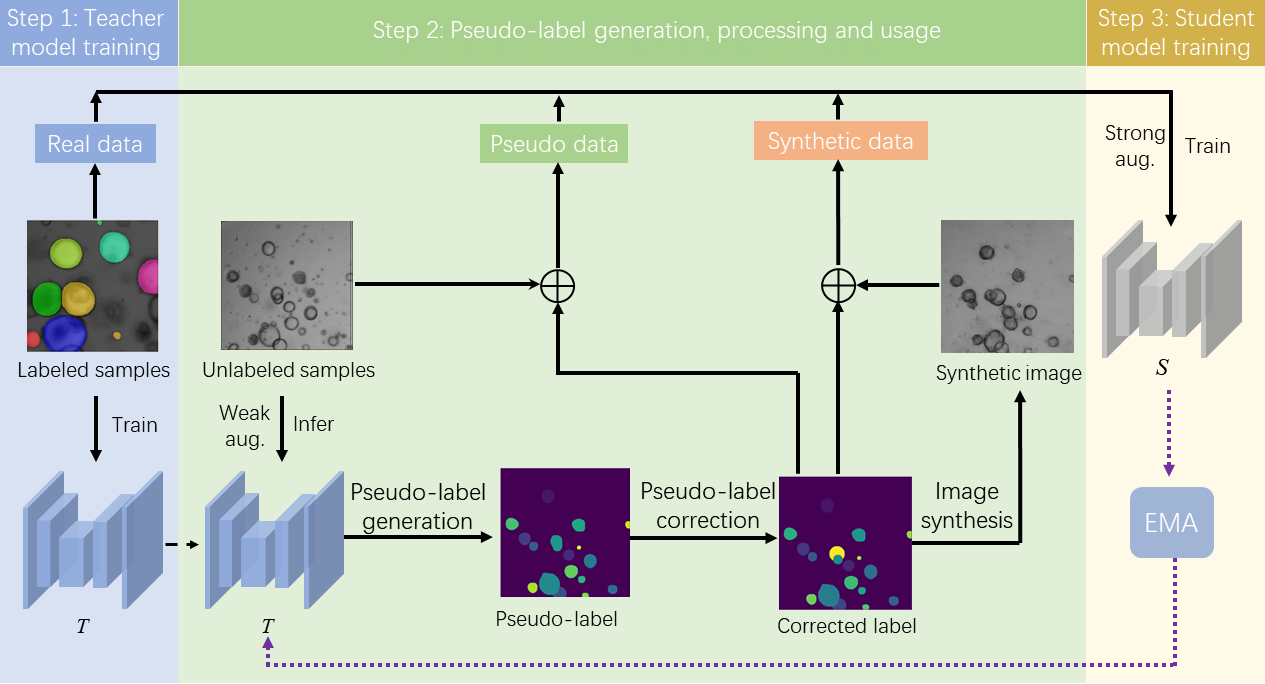}
	\caption{Framework of SA-SSL. A teacher model (\textit{T}) is initially trained on a limited set of labeled images. The trained teacher model generates pseudo-labels for unlabeled images. After the PLU process, new images are synthesized through pseudo-label guided image synthesis to augment the training dataset. A student model (\textit{S}) is trained on a combination of real data, pseudo-labeled data, and synthetic data. The weights of the trained \textit{S} are periodically updated to \textit{T} via EMA.}
	\label{fig4}
\end{figure}

\section{Experiments}
\subsection{Experimental Setup}
\subsubsection{Datasets}
The annotated datasets were developed as described in the methodology. For OrgaSegment, we adopt the training, validation, and testing set division established by Juliet \textit{et al} \cite{ref9}. For M-OrgaQuant, we follow the training and testing set division proposed by Kassis \textit{et al}. \cite{ref17} and randomly select 100 images from the original training set to constitute our validation set. As a result, the final dataset includes 1,540 unique images for training, 100 for validation, and 112 for testing, each containing at least one instance. The total number of instances are 17,937 for training, 915 for validation and 1,647 for testing, respectively. For semi-supervised training, we randomly sample 1\%, 2\%, 5\% and 10\% of the training set as the labeled training subset, with the remainder serving as the unlabeled images.

\subsubsection{Evaluation Metrics}
To thoroughly assess the performance of the trained instance segmentation model, four widely utilized evaluation metrics were adopted. Below is an introduction to these metrics:

1) mean of Average Precision (mAP): Calculated as the average of the areas under the precision-recall curves, then averaged across classes. It underscores the model's proficiency in detecting instances with both high precision and recall. 

2) F1-score (F1): The harmonic mean of precision and recall. It emphasizes the model's balance in mitigating false positives and false negatives, indicating its consistency in instance detection.

3) Dice coefficient (Dice): Measures the spatial overlap between predicted segmentations and ground truths. It assesses the model's accuracy in delineating the contours of individual instances, a crucial aspect for precise segmentation.

4) aggregated Jaccard index (AJI): AJI is an aggregated form of the Jaccard Index, evaluating the overlap between predicted segmentations and their corresponding ground truth. It captures both the number of correctly identified instances and the accuracy of their segmentations, providing a holistic view of the model's performance.

\subsubsection{Implementation Details}
The implementation is based on PyTorch 1.9.1, and all experiments are conducted on an NVIDIA RTX A6000 GPU. We employ Mask-RCNN as our baseline model, utilizing a ResNet-50 backbone with a Feature Pyramid Network (FPN) for feature extraction and multi-scale object detection. During training, we utilize SGD with a momentum of 0.9 as the optimizer. The initial learning rate is set to 0.001, with a linear warm-up applied for the first 1,000 iterations. The network undergoes 180,000 iterations, with the learning rate systematically reduced by a factor of 0.1 at the 80\% and 90\% stages of the total training iterations. The batch size is set at 6, which included 4 labeled images and 2 unlabeled images. For data augmentation of pseudo-labels, the default configurations are as follows: random shifts of 1 to 10 pixels in both the x and y directions, rotation angles ranging from 0 to 360°, and scaling factors between 0.9 and 1.1.

\subsection{Main Results}
\subsubsection{Efficiency Comparison: Contour-based vs Mask-based Approaches}
As shown in Table 1, we evaluate the efficiency of contoured-based and mask-based approaches on the OrgaSegment and M-OrgaQuant test datasets by measuring the synthesis time (\textit{t}) per image and FID between real and synthetic images. Values are presented as mean ± standard deviation. Smaller FID value indicates better alignment between the distributions of real and synthetic images. Our experiments demonstrate that the ‌contour-based approach‌ achieves higher synthesis efficiency, with significantly lower \textit{t} and FID values compared to the mask-based method. This performance advantage arises from: 1) reduced computational complexity, as mask-based approaches require multi-layer decomposition of overlapping structures, and 2) better feature preservation, where contour representations maintain more consistent distributions than decomposed sub-masks, resulting in lower FID scores.

\begin{table}[htbp]
	\caption{Efficiency assessments of contoured-based and mask-based approaches on the OrgaSegment and M-OrgaQuant test datasets\label{tab:table1}}
	\centering
	\setlength{\tabcolsep}{2pt} 
	\begin{tabular}{cccccc }
		\toprule[1pt]
		\multirow{2}{*}{Method} & \multicolumn{2}{c}{OrgaSegment} & & \multicolumn{2}{c}{M-OrgaQuant} \\ 
		\cmidrule{2-3} \cmidrule{5-6}
		& \textit{t} (s) & FID (\%) & & \textit{t} (s) & FID (\%) \\ 
		\midrule[0.5pt]
		Contour-based & 10.27 ± 6.16 & 137.3 ± 2.5 & & 2.72 ± 0.83 & 91.7 ± 0.9 \\ 
		Mask-based    & 21.06 ± 9.14 & 172.4 ± 3.7 & & 4.32 ± 1.51 & 95.5 ± 1.6 \\ 
		\bottomrule[1pt]
	\end{tabular}
\end{table}

\subsubsection{Effect of Augmentation Methods on SA-SSL}
In this part, we initially assess the impact of augmentation strategies for pseudo-labels on the FID between real and synthetic images. Results are listed in Table 2. Our experiments revealed that applying data augmentation to pseudo-labels resulted in an increase in FID. The most significant increase is observed when all augmentation techniques, including translation, rotation, and scaling, are applied simultaneously. Conversely, the smallest increase is noted when only scale augmentation is used. This suggests that while data augmentation enhances the diversity of training data, it also amplifies the distribution shift between real and synthetic images. Therefore, it is crucial to carefully select and combine augmentation techniques to achieve a balance between data diversity and distribution alignment.

\begin{table*}[!t]
	\centering
	\begin{threeparttable}
		\caption{FID between real and synthetic images before and after applying various data augmentation strategies\label{tab:table2}}
		\footnotesize 
		\setlength{\tabcolsep}{6pt} 
		\begin{tabular}{@{}ccccccccc@{}}
			\toprule[1pt]
			Augmentation & \multicolumn{4}{c}{OrgaSegment} & \multicolumn{4}{c}{M-OrgaQuant} \\
			\cmidrule(lr){2-5} \cmidrule(lr){6-9}
			& 1\% & 2\% & 5\% & 10\% & 1\% & 2\% & 5\% & 10\% \\
			\midrule
			Without & 159.4 ± 1.2 & 143.9 ± 0.7 & 123.8 ± 0.9 & 105.8 ± 1.0 & 106.5 ± 0.8 & 96.1 ± 0.5 & 82.7 ± 0.6 & 70.7 ± 0.7 \\
			S & 161.2 ± 1.2 & 154.6 ± 0.7 & 151.2 ± 1.2 & 115.4 ± 1.0 & 107.7 ± 0.8 & 103.3 ± 0.5 & 101.0 ± 0.8 & 77.1 ± 0.7 \\
			R & 168.0 ± 1.2 & 163.5 ± 0.9 & 160.0 ± 1.2 & 123.1 ± 1.2 & 112.2 ± 0.8 & 109.2 ± 0.6 & 106.9 ± 0.8 & 82.2 ± 0.8 \\
			T & 169.0 ± 1.3 & 165.4 ± 1.2 & 161.8 ± 1.2 & 141.3 ± 1.2 & 112.9 ± 0.9 & 110.5 ± 0.8 & 108.1 ± 0.8 & 94.4 ± 0.8 \\
			SR & 170.5 ± 1.2 & 166.2 ± 0.9 & 162.6 ± 1.2 & 126.0 ± 1.2 & 113.9 ± 0.8 & 111.0 ± 0.6 & 108.6 ± 0.8 & 84.2 ± 0.8 \\
			ST & 170.8 ± 1.3 & 172.5 ± 1.0 & 164.2 ± 1.3 & 142.1 ± 1.2 & 114.1 ± 0.9 & 115.2 ± 0.7 & 109.7 ± 0.9 & 94.9 ± 0.8 \\
			RT & 174.1 ± 1.3 & 172.5 ± 1.0 & 168.9 ± 1.3 & 141.5 ± 1.3 & 116.3 ± 0.9 & 115.2 ± 0.7 & 112.8 ± 0.9 & 94.5 ± 0.9 \\
			SRT & 175.7 ± 1.3 & 173.4 ± 1.0 & 170.5 ± 1.3 & 144.5 ± 1.3 & 117.4 ± 0.9 & 115.8 ± 0.7 & 113.9 ± 0.9 & 96.5 ± 0.9 \\
			\bottomrule[1pt]
		\end{tabular}
		\begin{tablenotes}
			\raggedright\footnotesize
			\item[] \textit{Abbreviations:} S: Scaling, R: Rotation, T: Translation, SR: Scaling+Rotation, ST: Scaling+Translation, RT: Rotation+Translation, SRT: Scaling+Rotation+Translation
		\end{tablenotes}
	\end{threeparttable}
\end{table*}

Next, we assess the performance of models trained using our framework with various augmentation strategies and compare these results to those obtained from models trained without any augmentation. As depicted in Figure 5, augmentation strategies that substantially increase FID had a mixed impact on model performance, improving certain metrics while potentially degrading others. Conversely, when scale augmentation is applied exclusively, models consistently show improvements in mAP without compromising other evaluation metrics. This indicates that scale augmentation effectively balances diversity and distribution alignment in SA-SSL, despite a slight increase in FID between synthetic and real images. Given its advantages, scale augmentation was exclusively adopted as the augmentation strategy in subsequent experiments.
\begin{figure*}
	\centering
	\includegraphics[width=0.9\linewidth]{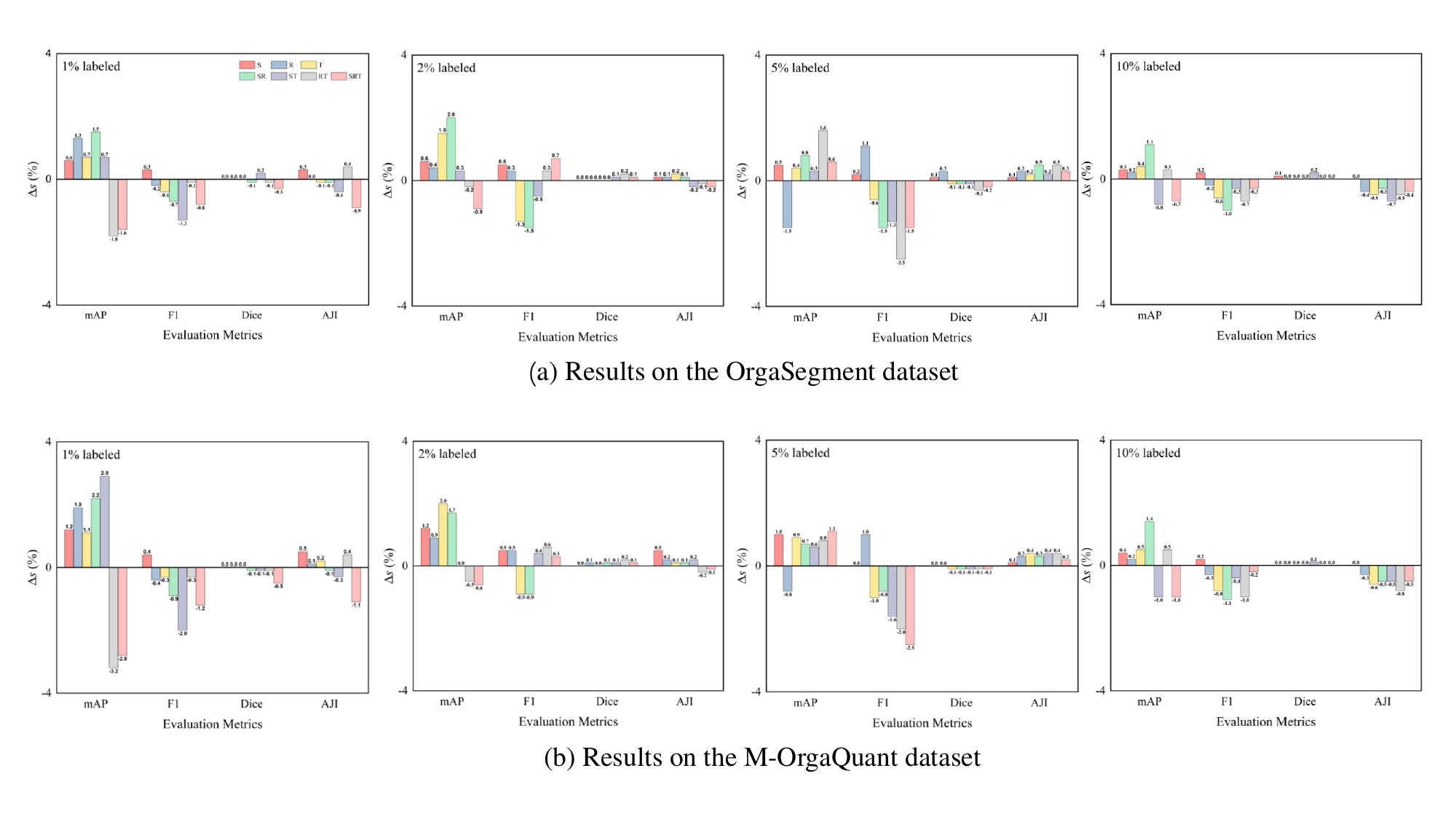}
	\caption{Changes in the evaluation metric scores ($\Delta S$) of the instance segmentation model after applying various data augmentation techniques under various labeled data conditions.}
	\label{fig5}
\end{figure*}

We further analyze the correlation between changes in a composite performance metric ($\Delta\mathrm{PM}$) and relative FID change ratios ($\Delta\mathrm{FID}$) induced by different augmentation strategies in Figure 6, aiming to gain preliminary insights into how to design augmentation strategies through FID analysis.$\Delta\mathrm{PM}$ is defined as the sum of scores from four model evaluation metrics. $\Delta\mathrm{FID}$ is calculated as:
\begin{equation}
	\Delta\mathrm{FID~}(\%)=(\mathrm{FID}_{A\mathrm{ug}}-\mathrm{FID}_{Au\mathrm{g}-free})/\mathrm{FID}_{Au\mathrm{g}-free}\times100 
	\label{eq}
\end{equation}
where $\mathrm{FID}_{A\mathrm{ug}}$ and $\mathrm{FID}_{Au\mathrm{g}-free}$  represent FID values with and without utilizing augmentation strategies, respectively. Our results demonstrate that increased $\Delta\mathrm{FID}$ generally correlates with decreased $\Delta\mathrm{PM}$and substantial FID increases lead to negative $\Delta\mathrm{PM}$ values, suggesting that optimal augmentation strategies should induce only marginal FID increases. While the precise correlations between $\Delta\mathrm{FID}$ and $\Delta\mathrm{PM}$ appears complex, we plan to systematically investigate this through expanded experiments with varied augmentation configurations and develop a predictive model in future work.

\begin{figure}
	\centering
	\includegraphics[width=0.9\linewidth]{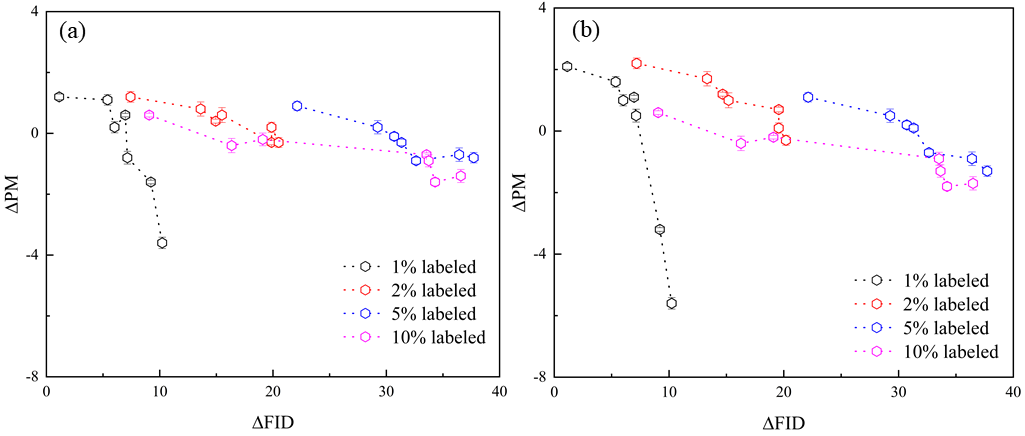}
	\caption{Variations in the composite model performance metric ($\Delta\mathrm{PM}$) and relative Fréchet Inception Distance (FID) change ratios ($\Delta\mathrm{FID}$) under different data augmentation strategies.(a) Results on the OrgaSegment dataset. (b) Results on the M-OrgaQuant dataset.}
	\label{fig6}
\end{figure}

\subsubsection{Comparisons Between SL and SSL Approaches}
In Table 3, we present a comparative analysis of the performance of supervised and SSL approaches, specifically SSL and our SA-SSL (Ours), on the test datasets. Notably, SSL consistently surpasses the supervised baseline in various evaluation metrics across different ratios of labeled data, except at the extremely low label rate of 1\% on the M-OrgaQuant dataset, where a declined mAP and AJI is observed. We hypothesize that this decrease is primarily caused by training bias resulting from a significant mismatch between the unlabeled images and their corresponding pseudo-labels, particularly for overlapping structures. This mismatch likely stems from the limited amount of labeled data, which restricts the teacher model's ability to generate high-quality pseudo-labels, thereby amplifying the noise introduced during training.

\begin{table*}[htbp]
	\centering
	\begin{threeparttable}
		\caption{ Comparisons of the model performance: supervised learning (SL) vs. semi-supervised learning (SSL) Approaches\label{tab:table3}}
		\setlength{\tabcolsep}{6pt}
        \scriptsize 
		\begin{tabular}{@{}cccccccccc@{}}
			\toprule[1pt]
			\multirow{2}{*} {Ratio} & \multirow{2}{*} {Method} & \multicolumn{4}{c}{OrgaSegment} & \multicolumn{4}{c}{M-OrgaQuant} \\
			\cmidrule(lr){3-6} \cmidrule(lr){7-10}
			& & mAP (\%) & F1 (\%) & Dice (\%) & AJI (\%) & mAP (\%) & F1 (\%) & Dice (\%) & AJI (\%) \\
			\midrule
			\multirow{3}{*}{1\%} 
			& SL & 32.8 ± 0.4 & 75.9 ± 0.8 & 55.1 ± 0.6 & 54.6 ± 0.5 & 74.2 ± 0.2 & 84.3 ± 0.3 & 70.5 ± 0.3 & 81.3 ± 0.2 \\
			& SSL & 39.1 ± 0.7 & 76.7 ± 0.8 & 57.4 ± 0.5 & 56.6 ± 0.2 & 73.5 ± 0.4 & 87.7 ± 0.4 & 70.3 ± 0.4 & 80.0 ± 0.1 \\
			& \textbf{Ours} & \textbf{41.6 ± 0.7} & \textbf{79.7 ± 0.9} & \textbf{59.8 ± 0.6} & \textbf{58.6 ± 0.3} & \textbf{79.7 ± 0.3} & \textbf{90.2 ± 0.4} & \textbf{78.2 ± 0.4} & \textbf{85.3 ± 0.1} \\
			\midrule
			\multirow{3}{*}{2\%}
			& SL & 35.5 ± 0.3 & 80.9 ± 0.5 & 59.6 ± 0.4 & 59.9 ± 0.1 & 74.3 ± 0.1 & 84.1 ± 0.3 & 71.5 ± 0.5 & 81.3 ± 0.1 \\
			& SSL & 38.7 ± 0.5 & 80.7 ± 0.3 & 60.4 ± 0.2 & 61.5 ± 0.3 & 76.7 ± 0.6 & 89.1 ± 0.1 & 76.4 ± 0.1 & 83.7 ± 0.2 \\
			& \textbf{Ours} & \textbf{41.6 ± 0.6} & \textbf{82.9 ± 0.3} & \textbf{61.9 ± 0.2} & \textbf{64.8 ± 0.2} & \textbf{80.6 ± 0.3} & \textbf{90.3 ± 0.1} & \textbf{78.6 ± 0.1} & \textbf{85.3 ± 0.1} \\
			\midrule
			\multirow{3}{*}{5\%}
			& SL & 42.1 ± 0.7 & 82.5 ± 0.4 & 60.3 ± 0.9 & 65.8 ± 0.3 & 76.9 ± 0.5 & 88.5 ± 0.7 & 75.2 ± 0.8 & 84.2 ± 0.2 \\
			& SSL & 43.3 ± 0.6 & 82.9 ± 0.6 & 61.5 ± 0.7 & 65.8 ± 0.2 & 78.1 ± 0.6 & 89.6 ± 0.2 & 77.5 ± 0.4 & 84.9 ± 0.1 \\
			& \textbf{Ours} & \textbf{45.6 ± 0.9} & \textbf{83.5 ± 0.7} & \textbf{63.7 ± 0.7} & \textbf{66.6 ± 0.3} & \textbf{81.7 ± 0.7} & \textbf{90.3 ± 0.4} & \textbf{79.3 ± 0.7} & \textbf{85.9 ± 0.1} \\
			\midrule
			\multirow{3}{*}{10\%}
			& SL & 43.5 ± 0.3 & 83.3 ± 0.4 & 62.6 ± 0.7 & 66.2 ± 0.5 & 77.8 ± 0.6 & 89.0 ± 0.6 & 77.8 ± 0.7 & 86.1 ± 0.1 \\
			& SSL & 45.7 ± 0.5 & 83.8 ± 0.8 & 64.2 ± 0.6 & 67.5 ± 0.3 & 80.5 ± 0.3 & 91.3 ± 0.4 & 80.1 ± 0.5 & 86.3 ± 0.1 \\
			& \textbf{Ours} & \textbf{47.6 ± 0.5} & \textbf{84.8 ± 0.7} & \textbf{65.7 ± 0.6} & \textbf{69.2 ± 0.4} & \textbf{82.8 ± 0.2} & \textbf{92.4 ± 0.4} & \textbf{81.6 ± 0.4} & \textbf{87.7 ± 0.1} \\
			\midrule
			100\% & SL & 47.6 ± 0.6 & 85.3 ± 0.9 & 66.1 ± 0.7 & 69.4 ± 1.0 & 82.5 ± 0.4 & 92.4 ± 0.6 & 81.8 ± 0.6 & 87.7 ± 0.5 \\
			\bottomrule[1pt]
		\end{tabular}
	\end{threeparttable}
\end{table*}

In contrast, our SA-SSL consistently outperforms both the supervised baseline and SSL, demonstrating its effectiveness in mitigating training bias and enhancing model performance. Meanwhile, as the amount of labeled data increases, both supervised and semi-supervised models exhibit significant performance improvements. Remarkably, our SA-SSL achieves results comparable to the fully supervised model, while using only 10\% of the labeled data. This demonstrates that our SA-SSL effectively fulfills the objectives of SSL by significantly reducing the reliance on extensive labeled datasets while maintaining competitive performance.

\subsubsection{Ablation Study for Our SA-SSL}

\begin{table*}[htbp]
	\centering
	\caption{Ablation study for our SA-SSL\label{tab:table4}}
	\setlength{\tabcolsep}{6pt}
    \scriptsize 
	\begin{threeparttable}
		\begin{tabular}{@{}ccccc*{4}{c}*{4}{c}@{}}
			\toprule[1pt]
			\multirow{2}{*}{Ratio} & \multirow{2}{*}{PD} & \multirow{2}{*}{PLU} & \multirow{2}{*}{SD} & \multirow{2}{*}{IA} & \multicolumn{4}{c}{OrgaSegment} & \multicolumn{4}{c}{M-OrgaQuant} \\
			\cmidrule(lr){6-9} \cmidrule(lr){10-13}
			&   &   &   &   & mAP (\%) & F1 (\%) & Dice (\%) & AJI (\%) & mAP (\%) & F1 (\%) & Dice (\%) & AJI (\%) \\
			\midrule
			\multirow{4}{*}{1\%} 
			& \checkmark & & & & 39.1±0.7 & 76.7±0.8 & 57.4±0.5 & 56.6±0.2 & 73.5±0.4 & 87.7±0.4 & 70.3±0.4 & 80.0±0.1 \\
			& \checkmark & \checkmark & & & 40.5±0.3 & 78.1±0.3 & 58.8±0.7 & 58.1±0.4 & 78.1±0.3 & 89.3±0.3 & 76.9±0.7 & 84.0±0.4 \\
			& \checkmark & \checkmark & \checkmark & & 41.0±0.3 & 79.4±0.2 & 59.8±0.3 & 58.3±0.1 & 78.5±0.3 & 89.8±0.2 & 78.2±0.3 & 84.8±0.1 \\
			& \checkmark & \checkmark & \checkmark & \checkmark & 41.6±0.7 & 79.7±0.9 & 59.8±0.6 & 58.6±0.3 & 79.7±0.3 & 90.2±0.4 & 78.2±0.4 & 85.3±0.1 \\
			\midrule
			\multirow{4}{*}{2\%}
			& \checkmark & & & & 39.7±0.5 & 80.7±0.3 & 60.4±0.2 & 61.5±0.3 & 76.7±0.6 & 89.1±0.1 & 76.4±0.1 & 83.7±0.2 \\
			& \checkmark & \checkmark & & & 41.0±0.1 & 81.1±0.4 & 61.2±0.5 & 62.7±0.3 & 79.3±0.1 & 89.5±0.4 & 77.4±0.5 & 84.3±0.3 \\
			& \checkmark & \checkmark & \checkmark & & 42.0±0.1 & 82.4±0.2 & 61.9±0.2 & 64.7±0.1 & 79.4±0.1 & 89.8±0.2 & 78.6±0.2 & 84.8±0.1 \\
			& \checkmark & \checkmark & \checkmark & \checkmark & 42.6±0.6 & 82.9±0.3 & 61.9±0.2 & 64.8±0.2 & 80.6±0.3 & 90.3±0.1 & 78.6±0.1 & 85.3±0.1 \\
			\midrule
			\multirow{4}{*}{5\%}
			& \checkmark & & & & 43.3±0.6 & 82.9±0.6 & 61.5±0.7 & 65.8±0.2 & 78.1±0.6 & 89.6±0.2 & 77.5±0.4 & 84.9±0.1 \\
			& \checkmark & \checkmark & & & 44.4±0.3 & 83.1±0.2 & 62.0±0.5 & 66.3±0.1 & 80.0±0.3 & 89.8±0.2 & 78.2±0.5 & 85.5±0.1 \\
			& \checkmark & \checkmark & \checkmark & & 45.1±0.2 & 83.3±0.2 & 63.8±0.4 & 66.5±0.1 & 80.7±0.2 & 90.3±0.2 & 79.3±0.4 & 85.8±0.1 \\
			& \checkmark & \checkmark & \checkmark & \checkmark & 45.6±0.9 & 83.5±0.7 & 63.7±0.7 & 66.6±0.3 & 81.7±0.7 & 90.3±0.4 & 79.3±0.7 & 85.9±0.1 \\
			\midrule
			\multirow{4}{*}{10\%}
			& \checkmark & & & & 45.7±0.5 & 83.8±0.8 & 64.2±0.6 & 67.5±0.3 & 80.5±0.3 & 91.3±0.4 & 80.1±0.5 & 86.3±0.1 \\
			& \checkmark & \checkmark & & & 46.1±0.5 & 84.2±0.2 & 65.1±0.2 & 68.4±0.1 & 81.2±0.5 & 91.7±0.2 & 81.2±0.2 & 87.4±0.1 \\
			& \checkmark & \checkmark & \checkmark & & 47.3±0.2 & 84.6±0.1 & 65.6±0.1 & 69.2±0.1 & 82.4±0.2 & 92.2±0.1 & 81.6±0.1 & 87.7±0.1 \\
			& \checkmark & \checkmark & \checkmark & \checkmark & 47.6±0.5 & 84.8±0.7 & 65.7±0.6 & 69.2±0.4 & 82.8±0.2 & 92.4±0.4 & 81.6±0.4 & 87.7±0.1 \\
			\bottomrule[1pt]
		\end{tabular}
	\end{threeparttable}
\end{table*}

\begin{table}
	\centering
	\caption{Comparison of performance on severely overlapping cses in the M-OrgaQuant dataset between models trained with the SA-SSL framework, without and with PLU\label{tab:table5}}
	\label{table}
	\setlength{\tabcolsep}{3pt}
	\begin{tabular}{c | c | c | c | c | c }
		\hline
		
		Ratio                 & PLU & mAP (\%) & F1 (\%) & Dice (\%)  & AJI (\%) \\
		\hline
		\multirow{2}{*}{1\%}                     &   & 39.5 ± 2.2 & 57.7 ± 1.3 & 43.5 ± 1.3  & 54.3 ± 1.2 \\ 
		& \checkmark  & 46.9 ± 1.4 & 61.5 ± 1.7 & 47.9 ± 0.9  & 56.4 ± 1.6 \\ 
		\hline                      
		\multirow{2}{*}{2\%}                     &   & 44.5 ± 1.6 & 57.7 ± 1.1 & 45.4 ± 1.1  & 54.9 ± 1.2 \\ 
		& \checkmark  & 50.4 ± 1.1 & 61.5 ± 1.3 & 54.3 ± 1.5  & 56.3 ± 1.1 \\ 
		\hline                 
		\multirow{2}{*}{5\%}  &             & 45.7 ± 1.5 & 60.0 ± 1.7 & 46.6 ± 1.8  & 55.2 ± 1.2 \\ 
		& \checkmark  & 51.9 ± 1.6 & 69.2 ± 1.2 & 54.3 ± 1.4  & 64.2 ± 1.1 \\ 
		\hline                     
		\multirow{2}{*}{10\%}                     &   & 48.6 ± 1.6 & 61.5 ± 1.6 & 49.7 ± 1.7  & 57.4 ± 1.1 \\ 
		& \checkmark  & 52.1 ± 1.3 & 69.2 ± 1.4 & 54.6 ± 1.5  & 64.6 ± 1.1 \\ 
		\hline   
		
	\end{tabular}
\end{table}

We perform ablation studies on our developed organoid datasets to investigate the effects of the different components in our proposed SA-SSL framework. The results are in Table 4. As detailed in the methodology, PLU serves as a vital preprocessing step for pseudo-labels, specifically designed to tackle the problem of incorrect pseudo-labels that arise in challenging scenarios. These scenarios typically involve organoids severely overlapping with others, leading to misidentification of multiple organoids as a single entity. Theoretically, PLU enhances the model performance by addressing the replacement of complex overlap in synthetic data with simpler and potentially misleading representations. This theoretical advantage is quantitatively validated in our comparative performance analysis, where PLU integration demonstrates significant improvements. We observe that the effect of PLU is more pronounced in the OrgaQuant data. We hypothesize that the limited gain in the OrgaSegment dataset stems from a scarcity of sufficiently challenging scenarios, constituting only 95 out of 11,962 instances ($\approx 0.8\%$) in OrgaSegment, compared to 1,968 out of 17,937 instances ($\approx 11.0\%$) in M-OrgaQuant. The roles of SD are examined by adopting it as a supplementary input to PD. Models trained with SD consistently outperform those without SD, highlighting that the integration of SD significantly enhances the performance of models trained with pseudo-labels. IA integration provides further performance enhancements.

To rigorously evaluate PLU's impact, we further assess model performance exclusively on challenging scenarios within the M-OrgaQuant dataset. The results (Table  5) demonstrate a significant quantitative improvement, underscoring PLU's effectiveness in enhancing the model's ability to handle complex scenarios.

\subsubsection{Comparison with State-of-the-Arts}

\begin{table*}[htbp]
	\centering
	\caption{Model performance comparison of diverse SSL methods\label{tab:table6}}
	\scriptsize 
	\begin{tabular}{c  c  c  c c c c  c c c c}
		\hline
		\multirow{2}{*}{Ratio} & \multirow{2}{*}{Methods} & \multirow{2}{*}{Backbones} & \multicolumn{4}{c}{OrgaSegment} & \multicolumn{4}{c}{M-OrgaQuant} \\
		\cmidrule(lr){4-7} \cmidrule(lr){8-11}
		& & & mAP (\%) & F1 (\%) & Dice (\%) & AJI (\%) & mAP (\%) & F1 (\%) & Dice (\%) & AJI (\%) \\
		\hline
		\multirow{6}{*}{1\%} 
		& ST & Mask-RCNN & 37.8 ± 0.7 & 74.8 ± 1.0 & 54.6 ± 0.7 & 54.7 ± 0.5 & 76.8 ± 0.8 & 88.1 ± 1.2 & 75.8 ± 0.9 & 82.7 ± 0.6 \\
		& NB & Mask-RCNN & 38.5 ± 0.5 & 75.1 ± 0.7 & 54.9 ± 0.5 & 55.0 ± 0.2 & 77.5 ± 0.4 & 87.8 ± 0.6 & 75.9 ± 0.7 & 82.7 ± 0.3 \\
		& PT & Center-Mask & 37.4 ± 0.6 & 74.8 ± 0.5 & 54.6 ± 0.3 & 54.7 ± 0.2 & 75.4 ± 0.5 & 87.1 ± 0.3 & 75.3 ± 0.6 & 82.4 ± 0.4 \\
		& PAIS & Mask-RCNN & 38.6 ± 0.7 & 75.5 ± 0.6 & 55.2 ± 0.4 & 55.8 ± 0.5 & 77.9 ± 0.5 & 88.0 ± 0.4 & 76.1 ± 0.6 & 84.0 ± 0.3 \\
		& GD & MaskFormer & 25.9 ± 1.8 & 61.2 ± 1.0 & 37.1 ± 2.1 & 42.9 ± 1.3 & 56.0 ± 2.2 & 72.3 ± 1.2 & 54.1 ± 2.5 & 66.8 ± 1.7 \\
		& \textbf{Ours} & \textbf{Mask-RCNN} & \textbf{41.6 ± 0.7} & \textbf{79.7 ± 0.9} & \textbf{59.8 ± 0.6} & \textbf{58.6 ± 0.3} & \textbf{79.7 ± 0.3} & \textbf{90.2 ± 0.4} & \textbf{78.2 ± 0.4} & \textbf{85.3 ± 0.1} \\
		\hline
		\multirow{6}{*}{2\%}
		& ST & Mask-RCNN & 38.2 ± 0.5 & 77.9 ± 1.0 & 56.9 ± 0.7 & 60.5 ± 0.8 & 77.7 ± 0.8 & 88.2 ± 1.2 & 76.2 ± 0.9 & 82.7 ± 0.6 \\
		& NB & Mask-RCNN & 39.0 ± 0.6 & 78.3 ± 0.4 & 57.2 ± 0.5 & 60.8 ± 0.5 & 78.4 ± 0.4 & 87.9 ± 0.6 & 76.3 ± 0.7 & 82.7 ± 0.3 \\
		& PT & Center-Mask & 37.8 ± 0.7 & 77.9 ± 0.5 & 56.8 ± 0.8 & 60.5 ± 0.3 & 76.3 ± 0.5 & 87.2 ± 0.3 & 75.7 ± 0.6 & 82.4 ± 0.4 \\
		& PAIS & Mask-RCNN & 39.0 ± 0.2 & 78.6 ± 0.6 & 57.4 ± 0.7 & 61.8 ± 0.5 & 78.7 ± 0.5 & 88.1 ± 0.4 & 76.5 ± 0.6 & 84.0 ± 0.3 \\
		& GD & MaskFormer & 26.3 ± 1.6 & 63.8 ± 1.1 & 38.8 ± 2.3 & 47.7 ± 1.4 & 56.7 ± 2.2 & 72.4 ± 1.2 & 54.4 ± 2.5 & 66.8 ± 1.7 \\
		& \textbf{Ours} & \textbf{Mask-RCNN} & \textbf{41.6 ± 0.6} & \textbf{82.9 ± 0.3} & \textbf{61.9 ± 0.2} & \textbf{64.8 ± 0.2} & \textbf{80.6 ± 0.3} & \textbf{90.3 ± 0.1} & \textbf{78.6 ± 0.1} & \textbf{85.3 ± 0.1} \\
		\hline
		\multirow{6}{*}{5\%}
		& ST & Mask-RCNN & 42.5 ± 0.5 & 78.5 ± 0.9 & 59.1 ± 0.8 & 62.6 ± 0.7 & 78.7 ± 0.8 & 88.2 ± 1.2 & 76.8 ± 0.9 & 83.2 ± 0.6 \\
		& NB & Mask-RCNN & 43.3 ± 0.6 & 78.8 ± 0.5 & 59.3 ± 0.8 & 62.9 ± 0.3 & 79.4 ± 0.4 & 87.9 ± 0.6 & 76.9 ± 0.7 & 83.2 ± 0.3 \\
		& PT & Center-Mask & 42.1 ± 0.4 & 78.5 ± 0.5 & 59.0 ± 0.7 & 62.6 ± 0.2 & 77.4 ± 0.5 & 87.2 ± 0.3 & 76.4 ± 0.6 & 83.0 ± 0.4 \\
		& PAIS & Mask-RCNN & 43.4 ± 0.4 & 79.2 ± 0.5 & 59.6 ± 0.7 & 63.9 ± 0.5 & 79.8 ± 0.5 & 88.1 ± 0.4 & 77.1 ± 0.6 & 84.6 ± 0.3 \\
		& GD & MaskFormer & 29.5 ± 1.5 & 64.2 ± 0.8 & 40.4 ± 1.2 & 49.4 ± 0.7 & 57.5 ± 2.2 & 72.4 ± 1.2 & 54.9 ± 2.5 & 67.3 ± 1.7 \\
		& \textbf{Ours} & \textbf{Mask-RCNN} & \textbf{45.6 ± 0.9} & \textbf{83.5 ± 0.7} & \textbf{63.7 ± 0.7} & \textbf{66.6 ± 0.3} & \textbf{81.7 ± 0.7} & \textbf{90.3 ± 0.4} & \textbf{79.3 ± 0.7} & \textbf{85.9 ± 0.1} \\
		\hline
		\multirow{6}{*}{10\%}
		& ST & Mask-RCNN & 45.5 ± 0.4 & 82.8 ± 0.7 & 63.5 ± 0.5 & 66.9 ± 0.4 & 80.6 ± 0.8 & 91.5 ± 1.2 & 80.0 ± 0.9 & 85.6 ± 0.6 \\
		& NB & Mask-RCNN & 46.1 ± 0.6 & 82.5 ± 0.8 & 63.6 ± 0.2 & 67.0 ± 0.5 & 80.9 ± 0.4 & 90.6 ± 0.6 & 79.9 ± 0.7 & 85.3 ± 0.3 \\
		& PT & Center-Mask & 44.9 ± 0.3 & 81.8 ± 0.5 & 63.2 ± 0.4 & 66.7 ± 0.2 & 78.9 ± 0.5 & 89.5 ± 0.3 & 79.2 ± 0.6 & 85.1 ± 0.4 \\
		& PAIS & Mask-RCNN & 46.3 ± 0.7 & 82.7 ± 0.2 & 63.8 ± 0.3 & 68.1 ± 0.5 & 81.4 ± 0.5 & 90.6 ± 0.4 & 80.0 ± 0.6 & 86.7 ± 0.3 \\
		& GD & MaskFormer & 32.6 ± 1.4 & 67.9 ± 0.5 & 45.1 ± 1.7 & 53.8 ± 0.8 & 60.5 ± 2.2 & 75.3 ± 1.2 & 59.1 ± 2.5 & 70.4 ± 1.7 \\
		& \textbf{Ours} & \textbf{Mask-RCNN} & \textbf{47.6 ± 0.5} & \textbf{84.8 ± 0.7} & \textbf{65.7 ± 0.6} & \textbf{69.2 ± 0.4} & \textbf{82.8 ± 0.2} & \textbf{92.4 ± 0.4} & \textbf{81.6 ± 0.4} & \textbf{87.7 ± 0.1} \\
		\hline
	\end{tabular}
\end{table*}

\begin{figure*}
	\centering
	\includegraphics[width=0.95\linewidth]{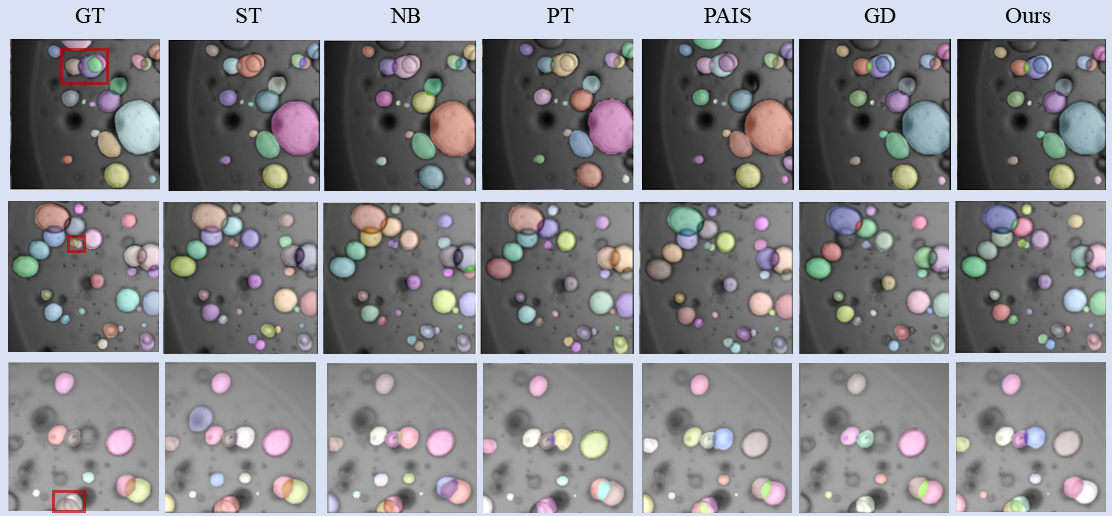}
	\caption{Instance segmentation results on the test datasets, comparing our method with semi-supervised baselines. Our approach achieves superior accuracy by effectively handling severely overlapping cases highlighted by red rectangles.}
	\label{fig7}
\end{figure*}

In this study, we compare our semi-supervised instance segmentation methods with the following state-of-the-art techniques:
\begin{itemize}
	\item Self-Training (ST) \cite{ref40}: presents a pseudo-labeling approach for SSL, where a teacher model trained on labeled data is utilized to generate pseudo-labels for unlabeled data, and then both the labeled and unlabeled data are employed to train a student model.
	\item Noisy Boundary (NB) \cite{ref23}: presents as an advanced semi-supervised instance segmentation model with advanced network architectures, where noise-tolerant mask head and boundary-preserving re-weighting are introduced to Mask-RCNN.
	\item Polite Teacher (PT) \cite{ref24}: is a semi-supervised instance segmentation model with anchor-free CNN detector, where pseudo-labels are refined by bounding boxes thresholding and mask scoring.
	\item PAIS \cite{ref25}: presents a pseudo-label refinement of Mask-RCNN to enhance semi-supervised instance segmentation, where pseudo-labels with low confidence but high mask quality are preferred to be kept.
	\item Guided Distillation (GD) \cite{ref26}: presents an advanced semi-supervised instance segmentation model training approach for transformed-based networks. Typically, it adopts MaskFormer, an anchor-free network, as the instance segmentation model.
\end{itemize}

Except for ST, the model training code for the other methods can be found at the corresponding links provided in the article. The training code for ST is accessible within the NB code repository. Models are trained using labeled data and evaluated on the test datasets. For a fair comparison, Resnet-50 was chosen as the backbone network for all CNN-based methods involved. The results of this comparison are presented in Table  6 and Figure 7. Notably, our methods outperform the state-of-the-art techniques across all evaluation metrics, demonstrating their effectiveness and robustness. Among these techniques, PAIS achieves optimal performance, primarily due to its mask scoring mechanism which partially mitigates segmentation challenges caused by severely overlapping instances. The relatively lower evaluation scores of Polite Teacher compared to other CNN-based methods might be attributed to the anchor-free network's sensitivity to overlap, which are prevalent in complex scenarios. The significant decline in model performance observed in Guided Distillation can be also ascribed to the heavy reliance of transformer-based networks on large amounts of data, indicating the advantage of CNN networks in achieving superior performance with limited labeled data.

\section{Conclusions}
In this study, we innovatively incorporate pseudo-label guided image synthesis, a technique previously utilized in the context of semi-supervised semantic segmentation, into the realm of SSL specifically tailored for organoid instance segmentation. This approach, combined with an efficient image synthesis method, a meticulously designed pseudo-label correction strategy, and a rigorously validated pseudo-label augmentation strategy, effectively enhances instance segmentation performance, even with a limited training dataset comprising merely a dozen images. Extensive experiments conducted on organoid segmentation datasets demonstrate that augmentation techniques that minimally reduce the structural similarity between real and synthetic images optimally enhance model performance, achieving a delicate balance between promoting diversity and minimizing distribution shifts. Our method presents a generic SSL framework that we believe will open up novel avenues for investigating label-efficient learning approaches in organoid applications, particularly in domains such as drug screening and personalized therapy. The rich information derived from individual organoids through instance segmentation models holds significant potential to facilitate these endeavors.

In future work, we aim to integrate adaptive instance augmentation techniques into our SA-SSL to better leverage synthetic data. This would further narrow the distribution shifts of SD. Additionally, we plan to evaluate our approach on alternative network architectures, such as Transformer-based models, to evaluate its generalizability across diverse model paradigms. Furthermore, we aim to develop adaptive augmentation techniques specifically tailored to organoid instance segmentation tasks, using extensive experiments and predictive modeling to optimize augmentation parameters and improve model performance and robustness.




\clearpage 




\section*{Declaration of competing interest}
The authors declare that they have no known competing financial interests or personal relationships that could have appeared to influence the work reported in this paper.

\section*{Funding sources}
This work was supported by the Joint Funds of the National Natural Science Foundation of China under Grant U24A20340, the National Natural Science Foundation of China under Grant 62471501, the Shenzhen Science and Technology Innovation Program under Grant JCYJ20220818102414031 and JCYJ20240813112459030, the Shenzhen Medical Research Fund under Grant B2402030,the Shenzhen Science and Technology Program JCYJ20220530150602006.

\section*{Data and code availability}
Data and code will be made available on request.




\end{document}